\documentclass[sigconf]{acmart}
\renewcommand\footnotetextcopyrightpermission[1]{} 
\usepackage[usenames,dvipsnames,svgnames,table]{xcolor}
\usepackage{xspace}
\usepackage[ruled,vlined]{algorithm2e}
\usepackage{multirow}
\usepackage{mathrsfs} 
\usepackage[normalem]{ulem} 
\usepackage{graphicx}
\usepackage{amsthm} 
\usepackage{amssymb}
\usepackage{amsmath}
\usepackage{fancyhdr}
\usepackage{subfig} 
\usepackage{url}
\usepackage{upgreek} 
\usepackage{tabularx}
\usepackage{booktabs}
\usepackage{changepage}
\usepackage{float}
\RequirePackage{comment}

\newcommand{\ignore}[1] {}

\newcommand{\lbl}[1] {\texttt{\small #1}}

\newcommand{\type}[1] {\lbl{#1}}
\newcommand{\instance}[1] {\textsc{#1}}

\newcommand{\universe}[1] {{\em {#1}}}

\newcommand{\removals}[1]{}

\newcommand{\done}[1]{}
\newcommand{\todoN}[1]{}

\newtheoremstyle{defistyling}
{\topsep}{\topsep}%
{}{}%
{\bfseries}{}
{\newline}
{%
  \thmname{#1}~\thmnumber{#2}\thmnote{\ -\ #3}.\\*[-1.5ex]%
}%

\newtheorem{definition}{Definition}[section]

\newtheorem{proposition}{Proposition}[section]

\newcommand{\tablebegin}[3]{
\begin{table}[!htb]
 \begin{center}
 \caption{#3}
 \label{#1}
\begin{tabular}{#2}
}

\newcommand{\tableend}{
\end{tabular}
\end{center}
\end{table}
}

\newcommand{\compactlist}{%
 \begin{list}{$\bullet$}{
    \setlength{\itemsep}{0pt}%
    \setlength{\parskip}{0pt}%
    \setlength{\parsep}{0pt}}}

\newcommand{\squishlist}{
 \begin{list}{$\bullet$}
  { \setlength{\itemsep}{2pt}
     \setlength{\parsep}{2pt}
     \setlength{\topsep}{2pt}
     \setlength{\partopsep}{0pt}
     \setlength{\leftmargin}{1.5em}
     \setlength{\labelwidth}{1em}
     \setlength{\labelsep}{0.5em} } }

\newcommand{\squishend}{\end{list}}

\renewcommand{\frac}[2]{\dfrac{#1}{#2}}

\newcommand{\simoncomment}[1]{\textit{\textcolor{green}{SR: #1}}}

\usepackage{booktabs} 
\usepackage{flushend}
\usepackage{placeins}
\usepackage{caption}
\usepackage{flushend}
\setcopyright{rightsretained}

	\copyrightyear{2019}
	\acmYear{2019}
	\acmConference[The Web Conference 2019]{The Web Conference 2019}{13--17 May, 2019}{San Francisco, USA}
	\acmPrice{}
	\acmDOI{}
	\acmISBN{}

\begin{document}

\title{TiFi: Taxonomy Induction for Fictional Domains\\ $[$Extended version$]$}
\titlenote{Conference version to appear at The Web conference, 2019.}


 \author{Cuong Xuan Chu}
 \affiliation{
     \institution{Max Planck Institute for Informatics}
     \city{Saarbr{\"u}cken}
     \country{Germany}
 }
 \email{cxchu@mpi-inf.mpg.de}

 \author{Simon Razniewski}
 \affiliation{
     \institution{Max Planck Institute for Informatics}
     \city{Saarbr{\"u}cken}
     \country{Germany}
 }
 \email{srazniew@mpi-inf.mpg.de}

 \author{Gerhard Weikum}
 \affiliation{
     \institution{Max Planck Institute for Informatics}
     \city{Saarbr{\"u}cken}
     \country{Germany}
 }
 \email{weikum@mpi-inf.mpg.de}



\begin{abstract}
Taxonomies are important building blocks of structured knowledge bases, and their construction from text sources and Wikipedia has received much attention. In this paper we focus on the construction of taxonomies for fictional domains, using noisy category systems from fan wikis or text extraction as input. Such fictional domains are archetypes of entity universes that are poorly covered by Wikipedia, such as also enterprise-specific knowledge bases or highly specialized verticals. Our fiction-targeted approach, called TiFi, consists of three phases: (i) category cleaning, by identifying candidate categories that truly represent classes in the domain of interest, (ii) edge cleaning, by selecting subcategory relationships that correspond to class subsumption, and (iii) top-level construction, by mapping classes onto a subset of high-level WordNet categories. A comprehensive evaluation shows that TiFi is able to construct taxonomies for a diverse range of fictional domains such as Lord of the Rings, The Simpsons or Greek Mythology 
with very high precision
and that it outperforms
state-of-the-art
baselines for taxonomy induction
by a substantial margin.
%
\end{abstract}

\maketitle
\thispagestyle{empty}


\section{Introduction}
\label{sec:introduction}
\subsection{Motivation and Problem}

\noindent{\bf Taxonomy Induction:}
Taxonomies, also known as type systems or class subsumption hierarchies, are an important resource for a variety of tasks related to text comprehension, such as information extraction, entity search or question answering. 
They represent structured knowledge about the subsumption of classes, for instance, that 
\type{electric guitar players} are \type{rock musicians}
and that \type{state governors} are \type{politicans}.
Taxonomies are a core piece of large knowledge graphs (KGs)
such as DBpedia, Wikidata, Yago and industrial KGs
at Google, Microsoft Bing, Amazon, etc.
When search engines receive user queries about
classes of entities, they can often find answers
by combining instances of taxonomic classes.
For example, a query about ``left-handed
electric guitar players'' can be answered
by intersecting the classes \type{left-handed people},
\type{guitar players} and \type{rock musicians};
a query about ``actors who became politicans''
can include instances from the intersection of
\type{state governors} and \type{movie stars}
such as Schwarzenegger.
Also, taxonomic class systems are very useful for
type-checking answer candidates for semantic search
and question answering \cite{DBLP:conf/semweb/KalyanpurMFW11}.

Taxonomies can be hand-crafted, examples being
WordNet~\cite{WordNet}, SUMO~\cite{DBLP:conf/fois/NilesP01} or
MeSH and UMLS~\cite{bodenreider2004unified}, or automatically constructed
by 
{\em taxonomy induction} from 
textual or semi-structured cues about 
type instances and subtype relations.
Methods for the latter include text mining 
using Hearst patterns~\cite{hearst}
or bootstrapped with Hearst patterns (e.g., \cite{DBLP:conf/sigmod/WuLWZ12}),
harvesting and learning from Wikipedia categories as
a noisy seed network (e.g., \cite{wikitaxo, wikitaxo2, wikitaxo3, yago, menta, WiBi, head, DBLP:conf/kdd/WuHW08}), 
and
inducing type hierarchies from query-and-click logs
(e.g., \cite{DBLP:conf/ijcai/PascaD07,DBLP:conf/emnlp/Pasca13,DBLP:journals/pvldb/GuptaHWWW14}).

\begin{figure}[b]
    \centering
    \vspace*{-0.5cm}
    \subfloat[LoTR]{{\includegraphics[scale=0.5]{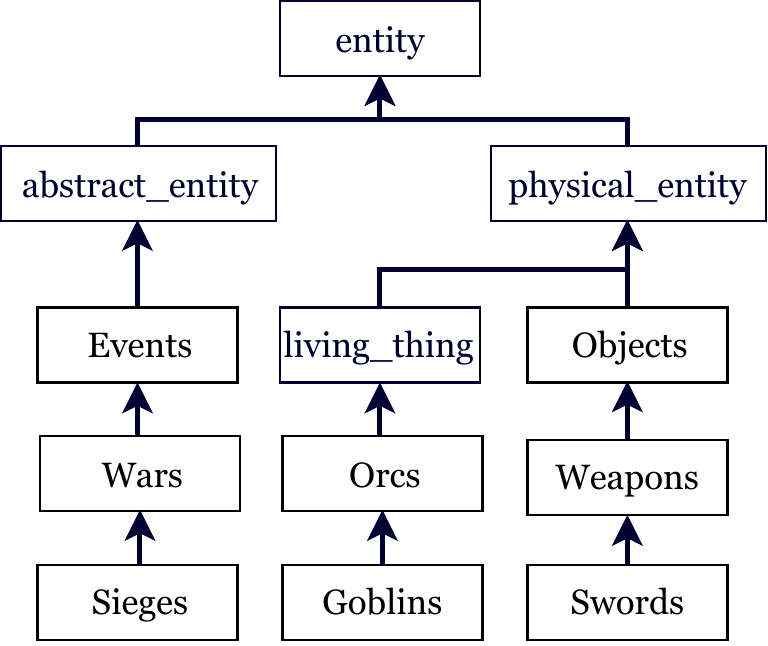}}}%
    \qquad
    \vspace*{-0.3cm}
    \subfloat[Star Wars]{{\includegraphics[scale=0.5]{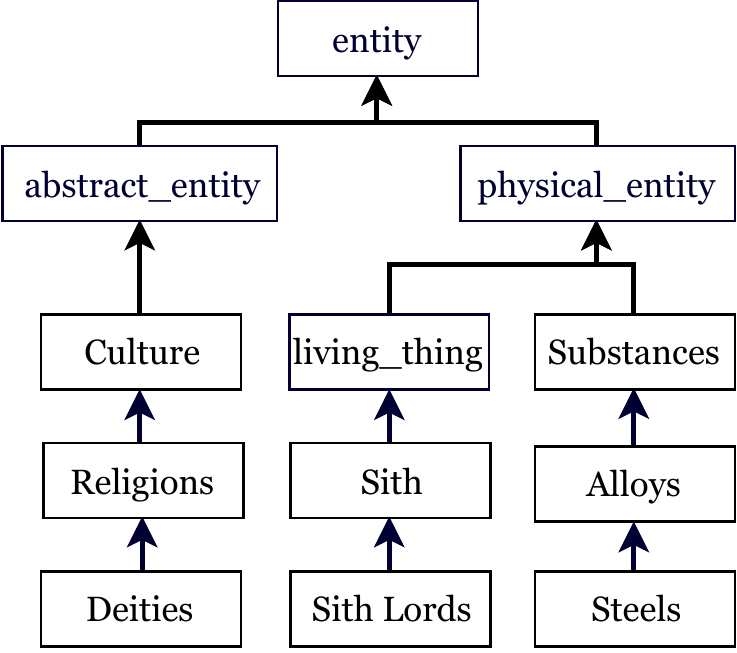} }}%
    \caption{Excerpts of LoTR and Star Wars taxonomies.}
    \label{fig:firstsample}
\end{figure}

\noindent{\bf The Case for Fictional Domains:} Fiction and fantasy are a core part of human culture, spanning from traditional literature to movies, TV series and video games. Well known fictional domains are, for instance,
the Greek mythology, the Mahabharata, 
Tolkien's Middle-earth,
the world of Harry Potter, or the Simpsons.
These universes contain many hundreds or even thousands
of entities and
types, and are subject of search-engine queries 
-- by fans as well as cultural analysts. 
For example, fans may query about Muggles who are 
students of the House of Gryffindor (within the
Harry Potter universe).
Analysts may be interested in understanding character relationships~\cite{feuding-families-NAACL16,bamman2014learning,srivastava2016inferring}, learning story patterns~\cite{chambers2009unsupervised,chaturvedi2017unsupervised} or investigating gender bias in different cultures~\cite{agarwal2015key}. 
Thus, organizing entities and classes from fictional domains into clean taxonomies (see example in Fig.~\ref{fig:firstsample}) is of
great value.

\noindent{\bf Challenges:} While taxonomy construction for encyclopedic
knowledge about the real world has received considerable attention already, taxonomy construction for fictional domains is a new problem that comes with 
specific challenges:
%
\squishlist
%
\item[1.] State-of-the-art methods for taxonomy induction make assumptions on 
entity-class and subclass relations that are often invalid for fictional domains.
For example, they assume that certain classes are disjoint (e.g., living beings 
and abstract entities, the oracle of Delphi being a counterexample).
Also, assumptions about the surface forms of entity names (e.g., on person names:
with or without first name, starting with Mr., Mrs., Dr., etc.)
and typical phrases for classes (e.g., noun phrases 
in plural form)
do not apply to fictional domains.
%
\item[2.] Prior methods for taxonomy induction intensively leveraged Wikipedia categories,
either as a content source or for distant supervision. 
However, the coverage of fiction and fantasy in Wikipedia is very limited,
and their categories are fairly ad-hoc. For example, 
 Lord Voldemort 
is in categories
like \type{Fictional cult leaders} (i.e., people), \type{J.K. Rowling characters} (i.e., a
meta-category) and \type{Narcissism in fiction}
(i.e., an abstraction). 
And whereas Harry Potter is reasonably covered in Wikipedia, fan websites feature many
more characters and domains such as House of Cards (a TV series) or
Hyperion Cantos (a 4-volume science fiction book)
that are hardly captured in Wikipedia.
%
\item[3.] Both Wikipedia and other content sources like fan-community forums
cover an ad-hoc mixture of in-domain and out-of-domain entities and types.
For example, they discuss both the fictional characters (e.g., Lord Voldemort)
and the actors of movies (e.g., Ralph Fiennes) and other aspects of
the film-making or book-writing.
\squishend

The same difficulties arise also when constructing
enterprise-specific taxonomies from highly
heterogeneous and noisy contents, or when
organizing types for highly specialized verticals
such as medieval history, 
the Maya culture, 
neurodegenerative diseases, 
or nano-technology material science.
Methodology for tackling such domains
is badly missing. 
We believe that our approach to fictional domains
has great potential for being carried over to
such real-life settings.
This paper focuses on fiction and fantasy, though,
where raw content sources are publicly available.


\subsection{Approach and Contribution}
 
In this paper we develop the first taxonomy construction method specifically geared
for fictional domains. We refer to our method as the {\bf TiFi} system, for
{\bf T}axonomy {\bf i}nduction for {\bf Fi}ction.
We address Challenge 1 by developing a classifier for categories and subcategory relationships 
that combines rule-based lexical and 
numerical contextual features. 
This technique is able to deal with difficult cases 
arising from non-standard entity names and class names.
Challenge 2 is addressed by tapping into fan community Wikis (e.g., {\small\tt harrypotter.wikia.com}).
This allows us to overcome the limitations of Wikipedia. 
Finally, Challenge 3 is addressed by constructing a supervised classifier for 
distinguishing in-domain vs. out-of-domain types, 
using a feature model specifically designed for fictional domains.

Moreover, we integrate our taxonomies with an upper-level taxonomy provided by WordNet, for
generalizations and abstract classes. This adds value for searching by entities and classes.
Our method outperforms the state-of-the-art taxonomy induction system for the first two steps, {HEAD}~\cite{head}, by 21-23\% and 6-8\%  percentage points in F1-score, respectively. An extrinsic evaluation based on entity search shows the value that can be derived from our taxonomies, where, for different queries, our taxonomies return answers with 24\% higher precision than the input category systems.
Along with the code of the TiFi system, we will 
publish taxonomies for 6 
fictional universes. 

\section{Related Work}

\paragraph{Text Analysis and Fiction}

Analysis and interpretation of fictional texts are an important part of cultural and language research, both for the intrinsic interest in understanding themes and creativity~\cite{chambers2009unsupervised,chaturvedi2017unsupervised}, and for extrinsic reasons such as predicting human behaviour~\cite{fast2016augur} or measuring discrimination~\cite{agarwal2015key}. Other recurrent topics are, for instance, to discover character relationships~\cite{feuding-families-NAACL16,bamman2014learning,srivastava2016inferring}, to model social networks~\cite{krishnan2014you,bamman2014learning}, or to describe personalities and emotions~\cite{elson2010extracting,jhavar-mirza-WWW18}.
Traditionally requiring extensive manual reading, automated NLP techniques have recently lead to the emergence of a new interdisciplinary subject called \emph{Digital Humanities}, which combines methodologies and techniques from sociology, linguistics and computational sciences towards the large-scale analysis of digital artifacts and heritage.

\paragraph{Taxonomy Induction from Text}
Taxonomies, that is, structured hierarchies of classes within a domain of interest, are a basic building block for knowledge organization and text processing, and crucially needed in tasks such as entity detection and linking, fact extraction, or question answering. A seminal contribution towards their automated construction was the discovery of Hearst patterns~\citep{hearst}, simple syntactic patterns like \emph{``X is a Y''} that achieve remarkable precision, and are conceptually still part of many advanced approaches. Subsequent works aim to automate the process of discovering useful patterns~\citep{snow2005learning, roller2016relations}. Recent work by Gupta et al.~\citep{gupta2017taxonomy} uses seed terms in combination with a probabilistic model to extract hypernym subsequences, which are then put into a directed graph from which the final taxonomy is induced by using a minimum cost flow algorithm. Other approaches utilize distributional representations of types~\cite{roller2014inclusive, yu2015learning,nguyen2017hierarchical, vu2018integrating}, or aim to learn them pairwise~\cite{yu2015learning} or hierarchically~\cite{nguyen2017hierarchical}. 

\begin{figure*}
 \centering 
 \includegraphics[scale=0.25]{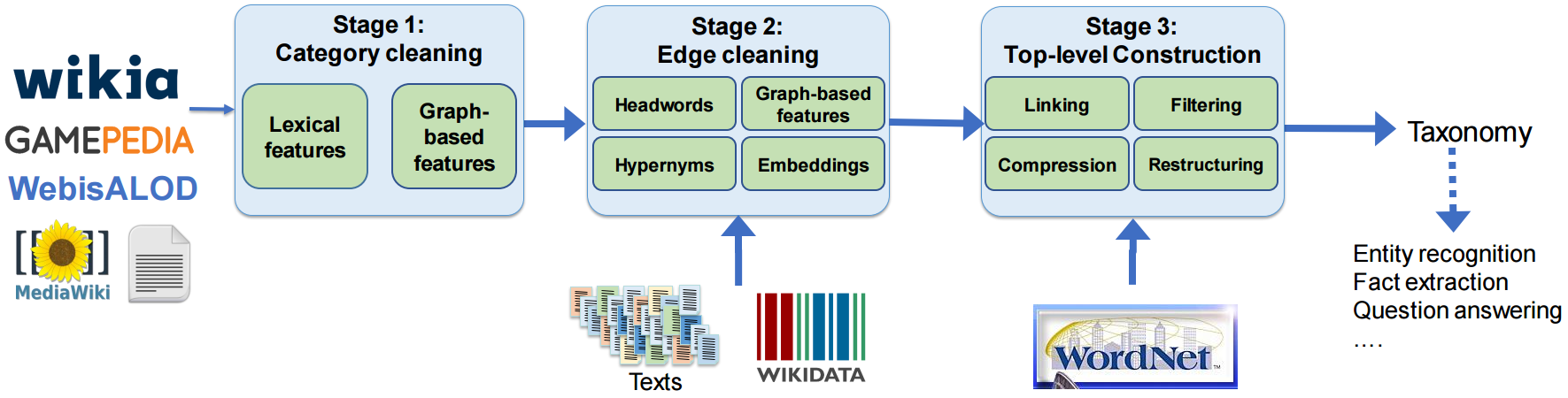}
  \caption{Architecture of TiFi.}
 \label{fig:architecture}
\end{figure*}

\paragraph{Taxonomy Construction using Wikipedia}

A popular structured source for taxonomy construction is the Wikipedia category network (WCN) for taxonomy induction. The WCN is a collaboratively constructed network of categories with many similarities to taxonomies, expressing for instance that the category \type{Italian 19th century composers} is a subcategory of \type{Italian Composers}. One project, WikiTaxonomy \cite{wikitaxo, wikitaxo3} aims to classify subcategory relations in the WCN as \textit{subclass} and \textit{not-subclass} relations. They investigate heuristics based on lexical matching between categories, lexico-syntactic patterns and the structure of the category network for that purpose. YAGO \cite{yago, yago2} uses a very simple criterion to decide whether a category represents a class, namely to check whether it is in plural form. It also provides linking to WordNet \cite{WordNet} categories, choosing in case of ambiguity simply the meaning appearing topmost in WordNet. MENTA~\cite{menta} learns a model to map Wikipedia categories to WordNet, with the goal of constructing a multilingual taxonomy over both. MENTA creates mean edges and subclass edges between categories and entities across languages, then uses Markov chains to rank edges and induce the final taxonomy. WiBi (Wikipedia Bitaxonomy)~\cite{WiBi} proceeds in two steps: It first builds a taxonomy from Wikipedia pages by extracting lemmas from the first sentence of pages, and heuristically disambiguating them and linking them to others. In the second step, WiBi combines the page taxonomy and the original Wikipedia category network to induce the final taxonomy. The most recent effort working on taxonomy induction over Wikipedia is HEAD \citep{head}. HEAD exploits multiple lexical and structural rules towards classifying subcategory relations, and is judiciously tailored towards high-quality extraction from the WCN.

\paragraph{Domain-specific Taxonomies}
TAXIFY is an unsupervised approach to domain-specific taxonomy construction from text~\cite{alfarone2015unsupervised}. Relying on distributional semantics, TAXIFY creates subclass  candidates, which in a second step are filtered based on a custom graph algorithm. Similarly, Liu et al.~\cite{liuetal} construct domain-specific taxonomies from keyword phrases augmented with relative knowledge and contexts. Compared with taxonomy construction from structured resources, these text-based approaches usually deliver comparably flat taxonomies.


\paragraph{Fan Wikis}

Fans are organizing content on fictional universes on a multitude of webspaces. Particularly relevant for our problem are fan Wikis, i.e., community-built web content constructed using generic Wiki frameworks. Some notable examples of such Wikis are  \textit{\url{tolkiengateway.net/wiki}}, with 12k articles, \textit{\url{www.mariowiki.com}} with 21k articles, or \textit{\url{en.brickimedia.org}} with 29k articles. Particularly relevant are also Wiki farms, like Wikia\footnote{\url{www.wikia.com/fandom}} and Gamepedia\footnote{\url{www.gamepedia.com}}, which host Wikis for 380k and 2k different fictional universes, and have Alexa rank 49 and 340, respectively.


In these Wikis, like on Wikipedia, editors collaboratively create and curate content. These Wikis come with support for categories, the \universe{The Lord of the Rings} Wiki, for instance, having over 900 categories and over 1000 subcategory relationships, the \universe{Star Wars} Wiki having 11k and 14k of each, respectively. Similarly as on Wikipedia, these category networks do not represent clean taxonomies in the ontological sense, containing for instance meta categories such as \type{1980 films}, or relations such as \type{Death\! in Battle} being a subcategory of \type{Character}.

\section{Design Rationale and Overview}

\subsection{Design Space and Choices}



\noindent{\em Input:} 
The input to the taxonomy induction problem is a set of entities, such as locations, characters and events, each with a description in the form of associated text or tags and categories. 
Entities with textual descriptions are easily available in many forums incl. Wikipedia, wikis of
fan communities or scholarly collaborations, and other online media. 
Tags and categories, including some form of category hierarchy, are available in various kinds of wikis
-- typically in very noisy form, though, with a fair amount of uninformative and misleading connections.
When such sites merely provide tags for entities, we can harness subsumptions between tags (e.g., simple association rules) to derive a {\em folksonomy} 
(see, e.g., \cite{DBLP:conf/esws/HothoJSS06,DBLP:conf/pkdd/JaschkeMHSS07,DBLP:journals/tmm/FangXSHG16}) 
and use this as an initial category system. When only text is available, we can use Hearst patterns and other text-based techniques \cite{hearst,DBLP:conf/sigir/SandersonC99,DBLP:journals/jair/CimianoHS05} to generate categories and construct a subsumption-based tree. 

\noindent{\em Output:}
Starting with a noisy category tree or graph for a given set of entities, from a domain of interest,
the goal of TiFi is to construct a clean taxonomy that preserves the valid and appropriate classes and their instance-of and subclass-of relationships but removes all invalid or misleading categories and connections. Formally, the output of TiFi is a directed acyclic graph (DAG) $G=(V,E)$ with 
vertices $V$ and edges $E$ such that 
(i) non-leaf vertices are semantic classes relevant for the domain,
(ii) leaf vertices are entities,
(iii) edges between leaves and their parents denote which entities belong to which classes,
(iv) edges among non-leaf vertices denote subclass-of relationships.

There is a wealth of prior literature on taxonomy induction methods, and the design space
for going about fictitious and other non-standard domains has many options.
Our design decisions are driven by three overarching considerations:
\squishlist
\item We leverage {\em whatever input information is available}, 
even if it comes with a high degree of noise.
That is, when an online community provides categories, we use them. 
When there are only tags or merely textual descriptions, we first build an initial category system using folksonomy construction methods and/or Hearst patterns.
\item For the output taxonomy, we {\em prioritize precision over recall}. So our methods mostly focus on removing invalid vertices and edges. Moreover, to make classes for fictitious domains
more interpretable and support cross-domain comparisons (e.g., for search), 
we aim to align the domain-specific classes with appropriate upper-level classes from
a general-purpose ontology, using WordNet \cite{WordNet}.
For example, dragons in Lord of the Rings should be linked to the proper WordNet sense
of dragons, which then tells us that this is a subclass of mythical creatures.
\item It may seem tempting to cast the problem into an end-to-end machine-learning task.
However, this would require sufficient training data 
in the form of pairs of input datasets and gold-standard output taxonomies. Such training data is not available,
and would be hard and expensive to acquire.
Instead, we break the overall task down into
focused steps at the granularity of individual
vertices and individual edges of category graphs.
At this level, it is much easier to acquire
labeled training data, by crowdsourcing (e.g., mturk).
Moreover, we can more easily devise features that
capture both local and global contexts,
and we can harness external assets like
dictionaries and embeddings.
\squishend

\subsection{TiFi Architecture}

\begin{figure*}
 \centering 
 \includegraphics[scale=0.67]{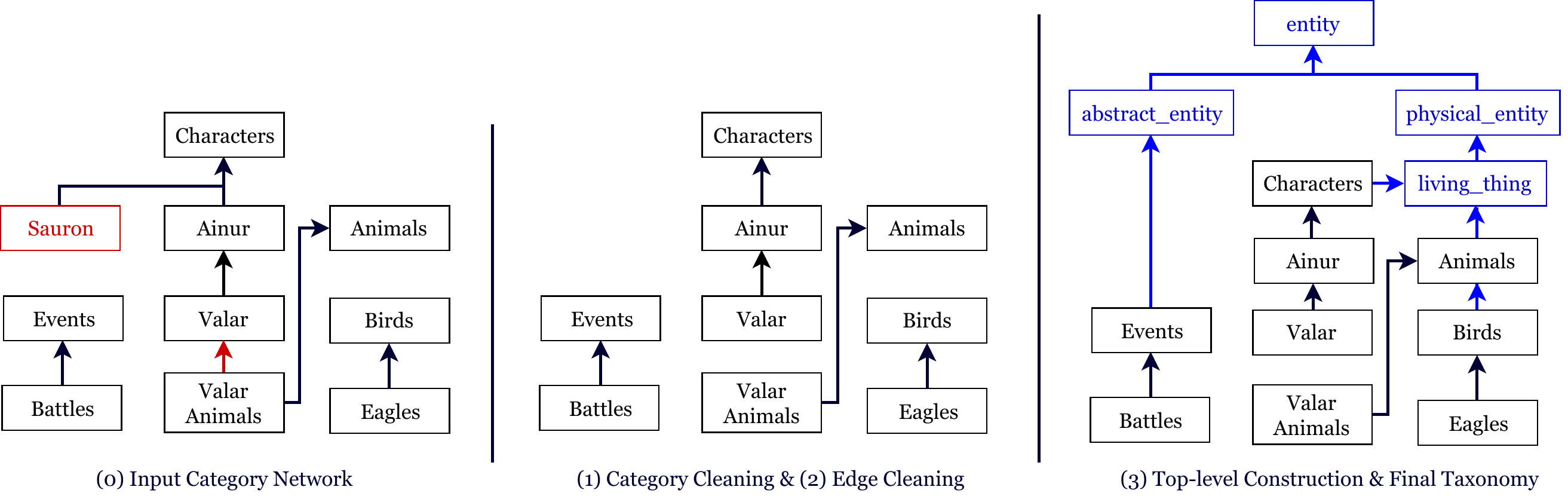}
 \caption{Example of three-stage taxonomy induction.} 
 \label{fig:taxonomyexample}
\end{figure*}

Based on the above considerations, we approach taxonomy induction in three steps, (1) category cleaning, (2) edge cleaning, (3) top-level construction.
The architecture of TiFi is depicted in Fig.~\ref{fig:architecture}.
Fig.~\ref{fig:taxonomyexample} illustrates how TiFi constructs a taxonomy.

The first step, \emph{category cleaning} (Section \ref{sec:noisycategorycleaning}), aims to clean the original set of categories $V$ by identifying categories that truly represent classes within the domain of interest, and by removing categories that represent, for instance, meta-categories used for community or Wikia coordination, or concern topics outside of the fictional domain, like movie or video game adaptions, award wins, and similar. Previous work has tackled this step via syntactic and lexical rules~\cite{yago,pasca2018finding,wikitaxo}. While such custom-tailored rules can achieve high accuracy, they have limitations w.r.t.\ applicability across domains. We thus opt for a supervised classification approach that combines rules from above with additional graph-based features. This way, taxonomy construction for a new domain only requires new training examples instead of new rules. Moreover, our experiments show that, to a reasonable extent, models can be reused across domains.

The second step, \emph{edge cleaning} (Section~\ref{sec:taxonomyinduction}), identifies the edges from the original category network $E$ that truly represent subcategory relationships. Here, both rule-based~\cite{DBLP:conf/coling/GuptaPKPP16,wikitaxo} and embedding-based approaches~\cite{nguyen2017hierarchical} appear in the literature. Each approach has its strength, however, rules again have limitations wrt.\ applicability across domains, while embeddings may disregard useful syntactic features, and crucially rely on enough textual content for learning. We thus again opt for a supervised approach, allowing us to combine existing lexical and embedding-based approaches with various adapted semantic and novel graph-based features.

For the third step, \emph{top-level construction} (Section \ref{sec:wordnetintegration}), basic choices are to aim to construct the top levels of taxonomies from input category networks~\cite{wikitaxo,DBLP:conf/coling/GuptaPKPP16}, or to reuse existing abstract taxonomies such as WordNet~\cite{yago}. As fan Wikis (and even Wikipedia) generally have a comparably small coverage of abstract classes, we here opt for the reuse of the existing WordNet top-level classes. This also comes with the additional advantage of establishing a shared vocabulary across domains, allowing to query, for instance, for \emph{animal species appearing both in LoTR and GoT}
(with answers such as dragons).

\section{Category Cleaning}
\label{sec:noisycategorycleaning}


In the first step, we aim to select the categories from the input that actually represent classes in the domain of interest. There are several reasons why a category would not satisfy this criterion, including the following:
\compactlist
    \item \textit{Meta-categories:} Wiki platforms typically introduce metacategories related to administration and technical setup, e.g., \type{Meta} or \type{Administration}.
    \item \textit{Contextual categories:} Community Wikis usually contain also information about the production of the universes (e.g., inspirations or actors), about the reception (e.g., awards), and about remakes and adaptions, which do not related to the real content of the universes.
    \item \textit{Instances:} Editors frequently create categories that are actually instances, e.g., \instance{Arda} or \instance{Mordor} in \universe{The Lord of The Rings}). 
    \item \textit{Extensions:} Wikis sometimes also contains fan-made extensions of universes that are not universally agreed upon. 
\squishend

Previous works on Wikipedia remove either only meta-categories or instances by using crafted lexical rules \cite{wikitaxo,wikitaxo3,pasca2018finding}. As our setting has to deal with a wider range of noise, we instead choose the use of supervised classification. We use a logistic regression classifier with binary (0/1) lexical and integer graph-based features, as detailed next.
\subsubsection*{\textbf{A. Lexical Features}}
\compactlist
\item \textit{Meta-categories}: True if a categories' name contains one of 22 manually selected strings, such as \type{wiki, template, user,} \type{portal,} \type{dis\-ambi\-guation, articles, administration, file,}\linebreak 
\type{pages}, etc.
\item \textit{Plural categories}: True if the headword of a category is in plural form. We use shallow parsing to extract headwords, for instance, identifying the plural term \type{Servants} in \type{Servants of Morgoth}, a strong indicator for a class.
\item \textit{Capitalization}: True if a category starts with a capital letter. We introduced this feature as we observed that in fiction, lowercase categories frequently represent non-classes.
\squishend

\subsubsection*{\textbf{B. Graph-based Features}}
\compactlist
\item \emph{Instance count:} The number of direct instances of a category.
\item \emph{Supercategory/\-sub\-category count:} The number of super/sub\-categories of a category, e.g., 0/2 for \type{Characters} in Fig.~\ref{fig:taxonomyexample} (left). Categories with more instances, superclasses or subclasses have potentially more relevance.
\item \emph{Average depth:} Average upward path length from a category. Categories with short paths above are potentially more likely not relevant. 
\item \emph{Connected subgraph size:} The maximal size of connected subgraphs which a given category belongs to. Each connected subgraph is extracted by using depth first search on each root of the input category network. Meta-categories are sometimes disconnected from the core classes of a universe.
\squishend
While the first two are established features, all other features have been newly designed to especially meet the characteristics of fiction. As we show in Section~\ref{sec:experiments}, this varied feature set allows to identify in-domain classes with 83\%-85\% precision.


\section{Edge Cleaning}
\label{sec:taxonomyinduction}

Once the categories that represent classes in the domain of interest have been identified, the next task is to identify which subcategory relationships also represent subclass relationships. While most previous works rely on rules~\cite{wikitaxo, menta, WiBi, head}, these are again too inflexible for the diversity of fictional universes. We thus tackle the task using supervised learning, relying on a combination of syntactic, semantic and graph-based features for a regression model.



\subsubsection*{\textbf{A. Syntatic Features}}

\paragraph{Head Word Matching}
Head word matching is arguably the most popular feature for taxonomy induction. Categories sharing the same headword, for instance \type{Realms} and \type{Dwarven Realms} are natural candidates for hypernym relationships.

\newcommand{\head}{\textit{head}}
\newcommand{\pre}{\textit{pre}}
\newcommand{\pos}{\textit{pos}}
\newcommand{\substr}{\textit{substr}}

We use a shallow parsing to extract, for a category $c$, its headword $\head(c)$, its prefix $\pre(c)$, and its suffix (postfix) $\pos(c)$, that is, $c = pre(c) + head(c) + pos(c)$. 
Consider a subcategory pair $(c_1,c_2)$:
\compactlist
\item[1.] If $\head(c_1) = \head(c_2)$, $\head(c_1) + \pos(c_1) = \head(c_2) + \pos(c_2)$ and $\pre(c_2) \subseteq \pre(c_1)$ then $c_2$ is a superclass of $c_1$.
\item[2.] If $\head(c_1) = \head(c_2)$, $\pre(c_1) + \head(c_1) = \pre(c_2) + \head(c_2)$ and $\pos(c_2) \subseteq \pos(c_1)$ then $c_2$ is a superclass of $c_1$.
\item[3.] If $\head(c_1) \neq \head(c_2)$ and $\head(c_2) \subseteq \pre(c_1)$ or $\head(c_2) \subseteq \pos(c_1)$ then there is no subclass relationship between $c_1$ and $c_2$.
\squishend
Case (1) covers the example of \type{Realms} and \type{Dwarven Realms}, while case (2) allows to infer, for instance, that \type{Elves} is a superclass of \type{Elves of  Gondolin}. Case (3) allows to infer that certain categories are not superclasses of each other, e.g., \type{Gondor} and \type{Lords of Gondor}.
Each of subclass and no-subclass inference are implemented as binary 0/1 features.

\paragraph{Only Plural Parent}
True if for a subcategory pair $(c_1,c_2)$, $c_1$ has no other parent categories, and $c_2$ is in plural form~\cite{head}.

\subsubsection*{\textbf{B. Semantic Features}}

\paragraph{{WordNet Hypernym Matching}}
WordNet is a carefully handcrafted lexical database that contains semantic relations between words and word senses (synsets), including hypo/hypernym relations. To leverage this resource, we map categories to WordNet synsets, using context-based similarity to identify the right word sense in the case of ambiguities. To compute the context vectors of categories, we extract their definitions, that is, the first sentence from the Wiki pages of the categories (if existing), and their parent and child class names. As context for WordNet synsets we use the definition (gloss) of each sense. We then compute cosine similarities over the resulting bags-of-words, and link each category with the position-adjusted most similar WordNet synset (see Alg.~\ref{alg:wnsynsetlinking}).
Then, given categories $c_1$ and $c_2$ with linked WordNet synset $s_1$ and $s_2$, respectively, this feature is true if $s_2$ is a WordNet hypernym of $s_1$.
\begin{algorithm}[!h]
 \KwData{A category $c$} 
 \KwResult{WordNet synset $s$ of $c$}
 $c = $ \textit{pre} + \textit{head} + \textit{pos}, $l$ = null\;
 $l$ = list of WordNet synset candidate for $c$\;
 \If{$l$ = null}{
    $l$ = list of WordNet synset candidates for \textit{pre} + \textit{head}\;
    \If{$l$ = null}{
        $l$ = list of WordNet synset candidates for \textit{head}\;
    }
 }
 \If{$l$ = null}{
    return null;
 }
 max = 0, $s$ = null\;
 \For{all WordNet synset $s_i$ in $l$}{
    $sim(s_i, c) = \textit{cosine}(V_{s_i}, V_c)$ with $V$: context vector\;
    $sim(s_i, c = sim(s_i, c) + 1/(2R_{s_i})$ where $R$: rank in WordNet\;
    \If{$\textit{sim}(s_i, c)$ > max}{
        max = $\textit{sim}(s_i, c)$\;
        $s$ = $s_i$\; 
    }
 }
 return $s$\;
 \caption{WordNet Synset Linking}
 \label{alg:wnsynsetlinking}
\end{algorithm}


\paragraph{{Wikidata Hypernym Matching}}
Similarly to WordNet, Wikidata also contains relations between entities. For example, Wikidata knows that \type{Maiar} is an instance (P31) of \type{Middle-earth races} in the \universe{The Lord of the Rings}. While Wikidata's coverage is generally lower than that of Wordnet, its content is sometimes complementary, as WordNet does not know certain concepts, e.g., \type{Maiar}.

\paragraph{{Page Type Matching}}
One interesting contribution of the WiBi system~\cite{WiBi} was to use the first sentence of Wikipedia pages to extract hypernyms. First sentences frequently define concepts, e.g., \emph{``The Haradrim, known in Westron as the Southrons and once as the ``Swertings'' by Hobbits, were a race of Men from Harad in the region of Middle-earth directly south of Gondor''}. For categories having matching articles in the Wikis, we rely on the first sentence from these. We use the Stanford Parser \cite{manning2014stanford} on the definition of the category to get a dependency tree. By extracting \texttt{nsubj, compound} and \texttt{conj} dependencies, we get a list of hypernyms for the category. For example, for \type{Haradrim} we can extract the relation \type{nsubj(race-13, Haradrim-2)}, hence \type{race} is a hypernym of \type{Haradrim}. After getting hypernyms for a category, we link these hypernyms to classes in the taxonomies by using head word matching, and set this feature to true for any pair of categories linked this way.

\paragraph{{WordNet Synset Description Type Matching}}
Similar to page type matching, we also extract superclass candidates from the description of the WordNet synset. For instance, given the WordNet description for \type{Werewolves}: \emph{``a monster able to change appearance from human to wolf and back again''}, we can identify \type{Monster} as superclass.

\paragraph{{Distributional Similarity}}
The distributional hypothesis states that similar words share similar contexts~\cite{harris1954distributional}, and despite the subclass relation being asymmetric, symmetric similarity measures have been found to be useful for taxonomy construction~\cite{shwartz2016hypernyms}. In this work, we utilize two distributional similarity measures, a symmetric one based on the structure of WordNet, and an asymmetric one based on word embeddings. The symmetric Wu-Palmer score compares the depth of two synsets (the headwords of the categories) with the depth of their least common subsumer ($\mathit{lcs})$~\cite{WuPalmer:ACL1994}. For synsets $s_1$ and $s_2$, it is computed as:   
\begin{equation} \label{eq:wnsim}
    \textit{Wu-Palmer}(s_1, s_2) = \frac{2 * \textit{depth}(lcs(s_1,s_2)) + 1}{\textit{depth}(s_1) + \textit{depth}(s_2) + 1}
\end{equation}

The HyperVec score~\cite{nguyen2017hierarchical} not only shows the similarity between a category and its hypernym, but is also directional. Given categories $c_1$ and $c_2$, with stemmed head words $h_1, h_2$ respectively, the HyperVec score is computed as:
\begin{equation} \label{eq:hypervec}
    \textit{HyperVec}(c_1, c_2) = \textit{cosine}(E_{h_1}, E_{h_2}) * \frac{||E_{h_2}||}{||E_{h_1}||},
\end{equation}
where $E_{h}$ is the embedding of word $h$. Specifically, we are using Word2Vec \cite{Mikolov:NIPS2013} to train a distributional representation over Wikia documents. The term $\textit{cosine}(E_{h_1}, E_{h_2})$ represents the cosine similarity between two embeddings, $||E_h||$ the Euclidean norm of an embedding. While WordNet only captures similarity between general concepts, embedding-based measures can cover both conceptual and non-conceptual categories, as often needed in the fantasy domain (e.g.\ similarity between \type{Valar} and \type{Maiar}).

\subsubsection*{\textbf{C. Graph-based Features}}

\paragraph{{Common Children Support}}
Absolute number of common children (categories and instances) of two given categories. Presumably, the more common children two categories have, the more related to each other they are.

\paragraph{{Children Depth Ratio}}
The ratio between the number of child categories of the parent of the edge, and its average depth in the taxonomy. This feature models the generality of the parent candidate.

\medskip

The features for edge cleaning combine existing state-of-the-art features (Head word matching, Page type matching, HyperVec) with adaptations specific to our domain (Wikidata hypernym matching, WordNet synset matching), and new graph-based features. Section~\ref{sec:experiments} shows that this feature set allows to surpass the state-of-the-art in edge cleaning by 6-8\% F1-score. 


\section{Top-level Construction}
\label{sec:wordnetintegration}

Category systems from Wiki sources often rather resemble forests than trees, i.e., do not reach towards very general classes, and miss useful generalizations such as \type{man-made structures} or \type{geographical features} for \type{fortresses} and \type{rivers}.
While works geared towards Wikipedia typically conclude with having identified classes and subclasses~\cite{wikitaxo, wikitaxo3, menta, WiBi, head}, we aim to include generalizations and abstract classes consistently across universes. For this purpose, TiFi employs as third step the integration of selected abstract WordNet classes. The integration proceeds in three steps:
\begin{enumerate}
\item Given the taxonomy constructed so far, nodes are linked to WordNet synsets using Algorithm \ref{alg:wnsynsetlinking}. Where the linking is successful, WordNet hypernyms are then added as superclasses. For example, the category \type{Birds} is linked to the WordNet synset \type{bird\%1:05:00::}, whose superclasses are \type{wn\_vertebrate $\rightarrow$ wn\_chordate $\rightarrow$ wn\_animal $\rightarrow$ wn\_organism $\rightarrow$ wn\_living\_thing $\rightarrow$ wn\_whole $\rightarrow$ wn\_object} $\rightarrow$ \ \ \ \ \ \  \linebreak \type{wn\_physical\_entity $\rightarrow$ wn\_entity}. 
\item The added classes are then compressed by removing those that have only a single parent and a single child, for instance, \type{abstract\_entity} and \type{physical\_entity} in Fig.~\ref{fig:taxonomyexample} (right) would be removed, if they really had only one child. 
\item We correct a few WordNet links that are not suited for the fictional domain, 
and use a self-built dictionary to remove 125 top-level WordNet synsets that are too abstract to add value, for instance, \type{whole}, \type{sphere} and \type{imagination}.
\end{enumerate}

Note that the present step can add subclass relationships between existing classes. In Fig.~\ref{fig:taxonomyexample}, after edge filtering, there is no relation between \type{Birds} and \type{Animals}, while after linking to WordNet, the subclass relation between \type{Birds} and \type{Animals} is added, making the resulting taxonomy more dense and useful.
\section{Evaluation}
\label{sec:experiments}

In this section we evaluate the performance of the individual steps of the TiFi approach, and the ability of the end-to-end system to build high-quality taxonomies.

\begin{table}[b]
\centering
\begin{tabular}{l|c|c}
\hline
\multicolumn{1}{c|}{\textbf{Universe}} & \textbf{\# Categories} & \textbf{\# Edges} \\ \hline
Lord of the Rings (LoTR)                                    &       973                 &      1118             \\ \hline
Game of Thrones (GoT)                        &       672                 &      1027             \\ \hline
Star Wars                                &       11012                 &     14092              \\ \hline
Simpsons                                &       2275                 &      4027             \\ \hline
World of Warcraft                       &       8249                 &      11403             \\ \hline
Greek Mythology                         &       601                 &       411            \\ \hline
\end{tabular}
\caption{Input categories from Wikia/Gamepedia.}
\label{tab:all-universe-statistics}
\vspace{-0.75cm}
\end{table}

\paragraph{Universes}

We use 6 universes that cover fantasy (LoTR, GoT), science fiction (Star Wars), animated sitcom (Simpsons), video games (World of Warcraft) and mythology (Greek Mythology). For each of these, we extract their category networks from dump files of Wikia or Gamepedia. The sizes of the respective category networks, the input to TiFi, are shown in Table \ref{tab:all-universe-statistics}.



\subsection{Step 1: Category Cleaning}
\label{sec:exp-noisycleaning}
Evaluation data for the first step was created using crowdsourcing, which was used to label all categories in LoTR, GoT, and random 50 from each of the other universes. 
Specifically, workers were asked to decide whether a given category had instances \emph{within} the fictional domain of interest. We collected three opinions per category, and chose majority labels. Worker agreement was between 85\% and 91\%.


As baselines we employ a rule-based approach by Ponzetto \& Strube~\cite{wikitaxo3}, to the best of our knowledge the best performing method for general category cleaning, and recent work by Marius Pasca~\citep{pasca2018finding} that targets the aspect of separating classes from instances. Furthermore, we combine both methods into a joint filter. The results of training and testing on LoTR/GoT, respectively, each under 10-fold crossvalidation, are shown in Table \ref{tab:noisycleaning-result-sameuniverse}. TiFi achieves both superior precision (+40\%) and F1-score (+22\%/+23\%), while observing a smaller drop in recall \mbox{(-18\%/-15\%)}. On both fully annotated universes the improvement of TiFi over the combined baseline in terms of F1-score is statistically significant  (p-value $2.2^{-16}$ and $1.9^{-13}$, respectively).
The considerable difference in precision is explained largely by the limited coverage of the rule-based baseline. 
Typical errors TiFi still makes are cases where categories have the potential to be relevant, yet appear to have no instances, e.g., \type{song} in LOTR. Also, it occasionally misses out on conceptual categories which do not have plural forms, e.g., \type{Food}.

\begin{table}
\centering
\scalebox{0.95}{
\begin{tabular}{c|l|c|c|c}
\hline
\textbf{Method}                                    & \multicolumn{1}{c|}{\textbf{Universe}} & \textbf{Precision} & \textbf{Recall} & \textbf{F1-score} \\ \hline

\multirow{2}{*}{\begin{tabular}[c]{@{}c@{}}Pasca \\ 2018 \cite{pasca2018finding}\end{tabular}} & LoTR          & 0.33      & 0.75    & 0.46    \\ \cline{2-5} 
                                          & GoT               & 0.57      & 0.85    & 0.68    \\ \hline 
                                          
\multirow{2}{*}{\begin{tabular}[c]{@{}c@{}}Ponzetto \& \\ Strube 2011 \cite{wikitaxo3}\end{tabular}} & LoTR          & 0.44      & \textbf{1.0}    & 0.61    \\ \cline{2-5} 
                                          & GoT               & 0.45      & \textbf{1.0}    & 0.62    \\ \hline 
                                          
\multirow{2}{*}{\begin{tabular}[c]{@{}c@{}}Pasca + \\ Ponzetto \& Strube\end{tabular}} & LoTR          & 0.41      & 0.75    & 0.53    \\ \cline{2-5} 
                                          & GoT               & 0.64      & 0.85    & 0.73    \\ \hline 
\multirow{2}{*}{TiFi}               & LoTR           & \textbf{0.84}      & 0.82   & \textbf{0.83}    \\ \cline{2-5} 
                                          & GoT               & \textbf{0.85}      & 0.85   & \textbf{0.85}    \\ \hline 
\end{tabular}}
\caption{Step 1 - In-domain category cleaning.}
\label{tab:noisycleaning-result-sameuniverse}
\vspace{-0.75cm}
\end{table}

\begin{table}[t]
\centering
\scalebox{1}{
\begin{tabular}{l|l|c|c|c}
\hline
\multicolumn{1}{c|}{\textbf{Train}} & \multicolumn{1}{c|}{\textbf{Test}} & \textbf{Precision}                                               & \textbf{Recall} & \textbf{F1-score} \\ \hline
LoTR                      & GoT                                & 0.81                                                             & 0.85            & 0.83             \\ \hline
GoT                                  & LoTR                    & 0.64                                                             & 0.88            & 0.74             \\ \hline
LoTR & Star Wars & 0.63 & 0.94 & 0.75 \\ \hline
LoTR & Simpsons & 0.91 & 0.63 & 0.74 \\ \hline
LoTR & World of Warcraft & 0.95 & 0.63 & 0.75 \\ \hline
LoTR & Greek Mythology & 0.86 & 0.6 & 0.71 \\ \hline
\end{tabular}}
\caption{Step 1 - Cross-domain category cleaning.}
\label{tab:noisycleaning-result-diffuniverse}
\vspace{-0.25cm}
\end{table}

A characteristic of fiction is variety. As our approach requires labeled training data, a question is to which extent labeled data from one domain can be used for cleaning categories of another domain. We thus next evaluate the performance when applying models trained on LoTR on the other 5 universes, and the model trained on GoT on LoTR. The results are shown in Table  \ref{tab:noisycleaning-result-diffuniverse}, where for universes other than LoTR and GoT, having annotated only 50 samples. As one can see, F1-scores drop by only 9\%/2\% compared with same-domain training, and the F1-score is above 70\% even for quite different domains.


To explore the contribution of each feature, we performed an ablation test using recursive feature elimination. The most important feature group were lexical features (30\%/10\% F1-score drop if removed in LoTR/GoT), with plural form checking being the single most important feature. In contrast, removing the graph-based features lead only to a 10\%/0\% drop, respectively.

\begin{table}
\centering
\scalebox{1}{
\begin{tabular}{l|l|c|c|c}
\hline
\textbf{Method}                      & \textbf{Universe} & \textbf{Precision} & \textbf{Recall} & \textbf{F1-score} \\ \hline

\multirow{2}{*}{HyperVec \cite{nguyen2017hierarchical}}       & LoTR     &  0.82               & 0.8            & 0.81  \\ \cline{2-5} 
                                     & GoT                  & \textbf{0.83}               & 0.81            & 0.82  \\ \hline 

\multirow{2}{*}{HEAD \cite{head}}       & LoTR     &  \textbf{0.85}               & 0.83            & 0.84  \\ \cline{2-5} 
                                     & GoT                  & 0.81               & 0.78            & 0.79  \\ \hline 
\multirow{2}{*}{TiFi} & LoTR                  & 0.83      & \textbf{0.98}   & \textbf{0.90} \\ \cline{2-5} 
                                     & GoT          & \textbf{0.83}      & \textbf{0.91}   & \textbf{0.87} \\ \hline 
\end{tabular}}
\caption{Step 2 - In-domain edge cleaning.}
\label{tab:taxinduc-result-sameuniverse}
\vspace{-0.25cm}
\end{table}

\begin{table}
\centering
\scalebox{0.9}{
\begin{tabular}{l|l|c|c|c|c}
\hline
\multicolumn{1}{c|}{\textbf{Train}} & \multicolumn{1}{c|}{\textbf{Test}} & \textbf{Precision}  & \textbf{Recall} & \textbf{F1-score} & \textbf{MAP} \\ \hline
LoTR & GoT & 0.81 & 0.79 & 0.80 & 0.92\\ \hline
GoT & LoTR & 0.89 & 0.87 & 0.88 & 0.89\\ \hline
GoT & Star Wars & 0.92 & 0.92 & 0.92 & 0.91\\ \hline
GoT & Simpsons & 0.86 & 0.86 & 0.86 & 0.92 \\ \hline
GoT & Word of Warcraft & 0.72 & 0.71 & 0.72 & 0.76 \\ \hline
GoT & Greek Mythology & 0.92 & 0.92 & 0.92 & 0.92 \\ \hline
\end{tabular}}
\caption{Step 2 - Cross-domain edge cleaning.}
\label{tab:taxinduc-result-diffuniverse-all}
\vspace{-0.75cm}
\end{table}

\begin{table*}[]
\centering
\scalebox{1}{
\begin{tabular}{c|l|c|c|c||c|c|c}
\hline
 &  &  \multicolumn{3}{c||}{\textbf{Proper-name edges}} & \multicolumn{3}{c}{\textbf{Concept edges}}\\
\hline
\textbf{Method}                      & \textbf{Universe} & \textbf{Precision} & \textbf{Recall} & \textbf{F1-score} & \textbf{Precision} & \textbf{Recall} & \textbf{F1-score} \\ \hline 

\multirow{2}{*}{HyperVec \cite{nguyen2017hierarchical}}       & LoTR     & 0.88                  & 0.59               & 0.71               & 0.80                  & 0.88               & 0.84               \\ \cline{2-8} 
                                     & GoT      & \textbf{1.0}                  & 0.16               & 0.27               & 0.83                  & 0.9                & 0.87               \\ \hline 
                                     
\multirow{2}{*}{HEAD \cite{head}}       & LoTR     & 0.91                  & 0.74               & 0.81               & 0.83                  & 0.87               & 0.85               \\ \cline{2-8} 
                                     & GoT      & 0.72                  & 0.68               & 0.70               & 0.82                  & 0.8                & 0.81               \\ \hline 
                                     
\multirow{2}{*}{TiFi} & LoTR     & \textbf{0.92}         & \textbf{0.79}      & \textbf{0.85}      & \textbf{0.88}         & \textbf{0.89}      & \textbf{0.88}      \\ \cline{2-8} 
                                     & GoT      & 0.96         & \textbf{0.68}      & \textbf{0.8}       & \textbf{0.90}         & \textbf{0.91}      & \textbf{0.91}      \\ \hline
\end{tabular}}
\caption{Step 2 - Edge cleaning: Proper-name vs.\ concept edges.}
\vspace{-0.5cm}

\label{tab:taxinduc-result-sameuniverse 2}
\end{table*}

\subsection{Step 2: Edge Cleaning}
\label{sec:exp-taxinduc}
We used crowdsourcing to label all edges that remained after cleaning noisy categories from LoTR, GoT, and random 100 edges in each of the other universes.
For example, we asked Turker whether in LOTR, \type{Uruk-hai} are \type{Orc Man Hybrids}. Inter-annotator agreement was between 90\% and 94\%.


We compare with two state-of-the-art systems: (1) HEAD~\cite{head}, the most recent system for Wikipedia category relationship cleaning, and (2) HyperVec \citep{nguyen2017hierarchical}, a recent embedding-based hypernym relationship learning system. The results for in-domain evaluation using 10-fold crossvalidation are shown in Table \ref{tab:taxinduc-result-sameuniverse}.
As one can see, TiFi achieves a comparable precision \mbox{(-2\%/+2\%),} and a superior recall (+15\%/+13\%), resulting in a gain in F1-score of 6\%/8\%. 
Again, the F1-score improvement of TiFi over HyperVec and HEAD on the two fully annotated universes is statistically significant (p-values $7.1^{-9}$, $0.01$, $5.8^{-5}$ and $6.5^{-5}$, respectively).

To explore the scalability of TiFi, we again perform cross-domain experiments using 100 labeled edges per universe. The results are shown in Table~\ref{tab:taxinduc-result-diffuniverse-all}. In all universes but \universe{World of Warcraft}, TiFi achieves more than 80\% F1-score, and the performance is further highlighted by mean average precision (MAP) scores above 89\%, meaning TiFi can effectively separate correct from incorrect edges.

As mentioned earlier, taxonomy induction on real-world domain can leverage a lot of semantic knowledge like WordNet synsets, while fiction frequently contains non-standard categories such as \type{Valar} and \type{Tatyar}. We further evaluate the performance of TiFi by distinguishing two types of edges:
\compactlist
\item \textit{Concept edges:} Both parent and child exist in WordNet.
\item \textit{Proper-name edges:} At least one of parent and child does not exist in WordNet.
\squishend
In \universe{The Lord of the Rings}, there are 145 proper-name edges and 407 concept edges, while in \universe{Game of Thrones}, there are 61 and 329 of each, respectively.
Table \ref{tab:taxinduc-result-sameuniverse 2} reports the performance of TiFi, comparing to HEAD and HyperVec on both types of edges. As one can see, for proper-name edges, TiFi achieves a very high precision of 92\%/96\%, outperforms HEAD by 4\%/10\% and HyperVec by 14\%/53\% in F1-score, respectively.

We again performed an ablation test in order to understand feature contribution.
We found that all three groups of features have importance, observing a 1-4\% drop in F1-score when removing any of them. The individually most important features were \textit{Only Plural Parent, Headword Matching, Common Children Support} and \textit{Page Type Matching}.

\subsection{Step 3: Top-level Construction}
\label{sec:exp-wn-integration}

The key step in top-level construction is the linking of categories to WordNet synsets (i.e. category disambiguation), hence we only evaluate this step. For this purpose, in each universe, we randomly selected 50 such links and evaluated their correctness, finding precisions between 84\% and 92\% (see Table \ref{tab:exp-wordnet-integration}). Overall, this step is able to link 30-72\% of top-level classes from Step 2, and adds between 22 to 373 WordNet classes and 76 to 3387 subclass relationships to our universes.

\begin{table}[t]
\centering
\scalebox{0.985}{
\begin{tabular}{l|c|c|c}
\hline
\textbf{Universe} & \textbf{\#New Types} & \textbf{\#New Edges} & \textbf{Precision} \\ \hline
LoTR     & 43                          & 171                       & 0.84                  \\ \hline
GoT      & 39                          & 179                       & 0.84                  \\ \hline
Starwars      &         373                  &       3387                 &      0.84             \\ \hline
Simpsons      &          115                 &       439                 &      0.92             \\ \hline
World of Warcraft      &     257                      &       2248                 &      0.84             \\ \hline
Greek Mythology      &      22                     &       76                 &        0.84           \\ \hline
\end{tabular}
} 
\caption{Step 3 - WordNet integration.}
\label{tab:exp-wordnet-integration}
\vspace{-0.75cm}
\end{table}

\begin{table}[b]
\centering
\begin{tabular}{l|c|c|c}
\hline
\textbf{Universe} & \textbf{\# Types} & \textbf{\# Edges} & \textbf{Precision} \\ \hline
LoTR              &         353          &          648         &     0.88               \\ \hline
Game of Thrones   &         292          &          497         &       0.83             \\ \hline
Star Wars         &         7352          &         12282          &     0.90               \\ \hline
Simpsons      &             1029      &             2171      &        0.88            \\ \hline
World of Warcraft            &      4063             &      7882             &        0.76            \\ \hline
Greek Mythology & 139 & 313 & 0.91 \\ \hline
\end{tabular}
\caption{Taxonomies produced by TiFi.}
\label{tab:fantasy-taxonomy-statistics}
\vspace{-0.5cm}
\end{table}

\subsection{Final Taxonomies}

\begin{figure*}
 \centering 
 \vspace{-0.25cm}
 \includegraphics[scale=0.44]{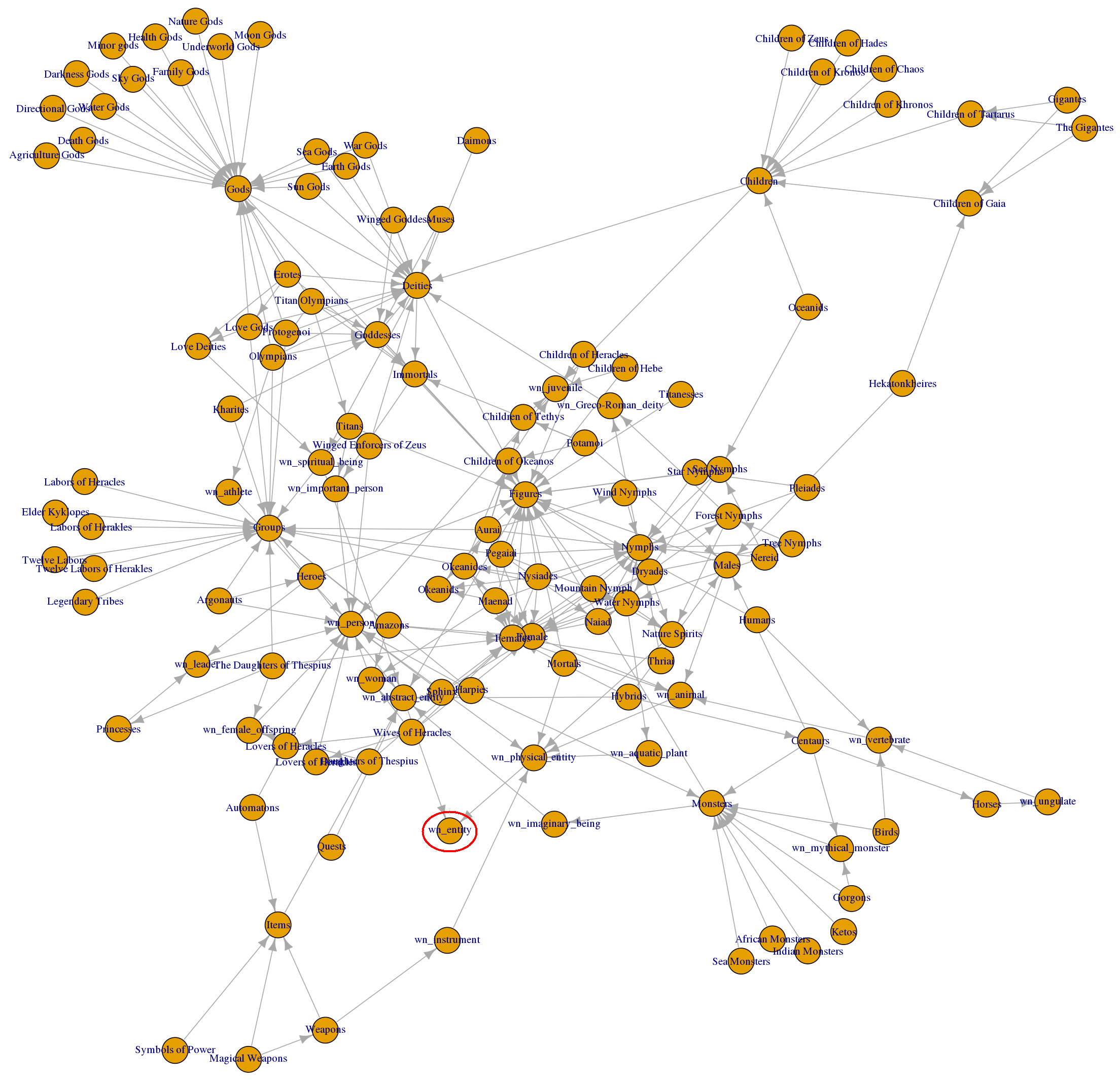}
\caption{Final TiFi taxonomy for Greek Mythology.}
 \label{fig:taxonomy-greek}
\end{figure*}

Table \ref{tab:fantasy-taxonomy-statistics} summarizes the taxonomies constructed for our 6 universes, with the bottom 4 universes built using the models for GoT. Reported precisions refer to the weighted average of the precision of subclass edges from Step 2, and the precision of WordNet linking. Figure~\ref{fig:taxonomy-greek} shows the resulting taxonomy for Greek Mythology, rendered using the R layout \emph{fruchterman.reingold}. All taxonomies will be made available both as CSV and graphically.

\subsection{Wikipedia as Input}

While our method is targeted towards fiction, it is also interesting to know how well it does in the traditional Wikipedia setting.
To this end, we extracted a specific slice of Wikipedia, namely all categories that are subcategories of \type{Desserts}, resulting in 198 categories connected by 246 subcategory relations, which we fully labeled.

Using 10-fold crossvalidation, in the first step, category cleaning, our method achieves 99\% precision and 99\% recall, which puts it on par with Ponzetto \& Strube \cite{wikitaxo3}, which achieves 99\% precision and 100\% recall. The reason for the excellent performance of both systems is that noise in Wikipedia categories concerns fairly uniformly meta-categories, which can be well filtered by enumerating them.
In the second step, edge cleaning, TiFi also achieves comparable results, with a slightly lower precision (83\% vs.\ 87\%) and a slightly higher recall (92\% vs.\ 89\%), resulting in 87\% F1-score for TiFi vs.\ 88\% for HEAD.

\subsection{WebIsALOD as Input}

WebIsALOD~\cite{webisalod} is a large collection of hypernymy relations extracted from the general web (Common Crawl). Relying largely on pattern-based extraction, the data from WebisALOD is very noisy, especially beyond the top-confidence ranks. Being text-based, several features based on category systems become unavailable, making this source an ideal stress test for the TiFi approach.

\paragraph{Data:} To get data from WebisALOD, we selected the top 100 most popular entities from two universes, \universe{The Lord of the Rings} and \universe{Simpsons}, 100 per each, based on the frequency of their mentions in text. We then queried the hypernyms of these entities and took the top 3 hypernyms based on ranking of confidences cores (minimum confidence 0.2). We iterated this procedure once with the newly gained hypernyms. In the end, with \universe{The Lord of the Rings}, we get 324 classes and 312 hypernym relations, meanwhile, with \universe{Simpsons}, these numbers are 271 classes and 228 hypernym relations. We fully manual label these datasets by checking whether classes are noisy and hypernym relations are wrong. From the labeled data, only 217 classes (67\%) and 167 classes (62\%) should be kept in \universe{The Lord of the Rings} and \universe{Simpsons}, respectively. In the case of hypernym relations, only 42\% and 47\% of them are considered to be correct relations in \universe{The Lord of the Rings} and \universe{Simpsons}, respectively. These statistics confirm that the data from WebisALOD is very noisy.

\paragraph{Results:} In Step 1, Ponzetto \& Strube \cite{wikitaxo3} use lexical rules to remove meta-categories, while Pasca \cite{pasca2018finding} uses heuristics which are based on information extracted from Wikipedia pages to detect entities that are classes. To enable comparison with Pasca's work, we used exact lexical matches to link classes from WebIsALOD to Wikipedia pages titles, then used Wikipedia pages as inputs. In fact, classes from WebisALOD are hardly meta-categories and the additional data from Wikipedia is also quite noisy. Table \ref{tab:noisycleaning-result-sameuniverse-webisA} shows that TiFi still performs very well in category cleaning, and significantly outperforms the baselines by 10\%/20\% F1-score. 

\begin{table}
\centering
\scalebox{1}{
\begin{tabular}{c|l|c|c|c}
\hline
\textbf{Method}                                    & \multicolumn{1}{c|}{\textbf{Universe}} & \textbf{Precision} & \textbf{Recall} & \textbf{F1-score} \\ \hline

\multirow{2}{*}{\begin{tabular}[c]{@{}c@{}}Pasca, \\ 2018 \cite{pasca2018finding}\end{tabular}} & LoTR          & 0.67      & \textbf{1.0}    & 0.80    \\ \cline{2-5} 
                                          & Simpsons               & 0.62   &   \textbf{1.0}    & 0.76    \\ \hline 
                                          
\multirow{2}{*}{\begin{tabular}[c]{@{}c@{}}Ponzetto \& \\ Strube 2011 \cite{wikitaxo3}\end{tabular}} & LoTR          & 0.67      & \textbf{1.0}    & 0.80    \\ \cline{2-5} 
                                          & Simpsons               & 0.62   &   \textbf{1.0}    & 0.76    \\ \hline 
                    
\multirow{2}{*}{TiFi}               & LoTR           & \textbf{0.89}      & 0.94   & \textbf{0.91}    \\ \cline{2-5} 
                                          & Simpsons               & \textbf{0.95}      & 0.97   & \textbf{0.96}    \\ \hline 
\end{tabular}}
\caption{WebIsALOD input - step 1 - In-domain cat.\ cleaning.}
\label{tab:noisycleaning-result-sameuniverse-webisA}
\vspace{-0.5cm}
\end{table}

\begin{table*}[h!]
\centering
\scalebox{0.69}{
\begin{tabular}{l|c|c|l|l}
\hline
\textbf{}              & \multicolumn{2}{c|}{\textbf{Text}}                                         & \multicolumn{2}{c}{\textbf{Structured Sources}}                                  \\ \hline
\multicolumn{1}{c|}{\textbf{Query}}                                                       & \textbf{Google}                                                                                                      & \textbf{Wikia}                                                                                  & \multicolumn{1}{c|}{\textbf{Wikia-categories}}                                                                                                                                                                                                                                                                                                                                                                                    & \multicolumn{1}{c}{\textbf{TiFi}}                                                                                                                                                                                                                                        \\ \hline
Dragons in LOTR                                                                            & \multicolumn{1}{l|}{\begin{tabular}[c]{@{}l@{}}Glaurung, \\ T\'{u}rin, Turambar, \\ E\"{a}rendil, Smaug,\\ Ancalagon\end{tabular}} & \multicolumn{1}{l|}{\begin{tabular}[c]{@{}l@{}}Dragons,\\ \sout{Summoned} \sout{Dragon},\\ Spark-dragons\end{tabular}} & \begin{tabular}[c]{@{}l@{}}\sout{Urgost},Long-worms,Gostir,\sout{Drogoth the Dragon Lord},\sout{Cave-Drake},\\ \sout{War of the Dwarves and Dragons}, \sout{Dragon-spell},Stone Dragons,\\ Fire-drake of Gondolin,Spark-dragons, Were-worms, \sout{Summoned}\\ \sout{Dragon}, Fire-drakes, Glaurung,Ancalagon,Dragons,Cold-drakes, \\ Sea-serpents, \sout{User blog:Alex Lioce/Kaltdrache the Dragon}, Smaug,\\  \sout{Dragon (Games}, \sout{Workshop)}, \sout{Drake}, Scatha, \sout{The Fall of Erebor}\end{tabular} & \begin{tabular}[c]{@{}l@{}}Long-worms, \sout{War of the Dwarves and Dragons},\\ \sout{Dragon-spell},Stone Dragons, Fire-drake of Gondolin,\\ Spark-dragons, Were-worms, Fire-drakes, Glaurung, \\ Ancalagon, Dragons, Cold-drakes, Sea-serpents,\\ Smaug, Scatha ,\sout{The Fall of Erebor}, Gostir \end{tabular} \\ \hline
\begin{tabular}[c]{@{}l@{}}Which Black Numenoreans \\ are servants of Morgoth\end{tabular} & -                                                                                                                    & Black N\'{u}men\'{o}rean                                                                                               & \begin{tabular}[c]{@{}l@{}}Men of Carn D\^{u}m,Corsairs of Umbar,Witch-king of Angmar,\\ \sout{Thrall Master},Mouth of Sauron,Black N\'{u}men\'{o}rean,Fuinur\end{tabular}                                                                                                                                                                                                                                                                & \begin{tabular}[c]{@{}l@{}}Men of Carn D\^{u}m,Corsairs of Umbar,Witch-king of \\ Angmar, Mouth of Sauron, Black N\'{u}men\'{o}rean, Fuinur\end{tabular}                                                                                                                                         \\ \hline
\begin{tabular}[c]{@{}l@{}}Which spiders \\ are not agents of Saruman?\end{tabular}        & -                                                                                                                    & -                                                                                                  & \begin{tabular}[c]{@{}l@{}}Shelob, \sout{Spider Queen and Swarm},\sout{Saenathra},\\\sout{Spiderling}, Great Spiders, \sout{Wicked, Wild, and Wrath}\end{tabular}                                            & Shelob, Great Spiders    \\ \hline
\end{tabular}}
\caption*{Table 12. Example queries and results for the entity search evaluation.}
\label{tab:entitysearch}
\vspace{-0.25cm}
\end{table*}


In Step 2, HEAD uses heuristics to clean hypernym relations between classes, mostly based on lexical and information from class pages (e.g. Wikipedia pages). Although TiFi also uses the information from class pages, its supervised model uses also a set of other features and is thus more versatile. Table \ref{tab:taxinduc-result-sameuniverse-webisA} reports the results of TiFi, comparing with HEAD in edge cleaning, with TiFi outperforming HEAD by 28\%-53\% F1-score. 

Both steps were also evaluated in the cross-domain settings, with similar results (90\%/91\% F1-score in step 1, 53\%/55\% F1-score in step 2).

\begin{table}
\centering
\scalebox{1}{
\begin{tabular}{l|l|c|c|c}
\hline
\textbf{Method}                      & \textbf{Universe} & \textbf{Precision} & \textbf{Recall} & \textbf{F1-score} \\ \hline

\multirow{2}{*}{HEAD \cite{head}}       & LoTR     &  0.27              & 0.05            & 0.09  \\ \cline{2-5} 
                                     & Simpsons                  & 0.31               & 0.09            & 0.14  \\ \hline 
\multirow{2}{*}{TiFi} & LoTR                  & \textbf{0.79}      & \textbf{0.55}   & \textbf{0.62} \\ \cline{2-5} 
                                     & Simpsons          & \textbf{0.61}      & \textbf{0.32}   & \textbf{0.42} \\ \hline 
\end{tabular}}
\caption{WebIsALOD - step 2 - In-domain edge cleaning.}
\label{tab:taxinduc-result-sameuniverse-webisA}
\vspace{-0.25cm}
\end{table}


\section{Use Case: Entity Search}
\label{sec:extrinsiceval}
To highlight the usefulness of our taxonomies, we provide an extrinsic evaluation based on the use case of entity search. Entity search is a standard problem in information retrieval, where often, textual queries shall return lists of matching entities. In the following, we focus on the retrieval of correct entities only, and disregard the ranking aspect.

\paragraph{Setup}
We consider three universes, \universe{The Lord of the Rings, Simpsons} and \universe{Greek Mythology}, and manually generated 90 text queries belonging to the following categories (10 of each per universe):
\begin{enumerate}
\item Single type: Entities belonging to a class, e.g., \textit{Orcs in the Lords of the Rings};
\item Type intersection: Entities belonging to two classes, e.g., \textit{Humans that are agents of Saruman};
\item Type difference: Entities that belong to one class but not another, e.g., \textit{Spiders that are not servants of Sauron}.
\end{enumerate}

We utilize the following resources:


\begin{itemize}
    \item Unstructured resources: (1) Google Web Search and (2) the Wikia-internal text search function;
    \item Structured resources: (3) the Wikia category networks and (4) the taxonomies as built by TiFi.
\end{itemize}

\paragraph{{Evaluation}} For the unstructured resources, we manually checked the titles of the top 10 returned pages for correctness.

For the structured resources, we matched the classes in the query against all classes in the taxonomy that contained those class names as substrings. We then computed, in a breadth-first manner, all subclasses and all instances of these classes, truncating the latter to maximal 10 answers, and manually verified whether returned instances were correct or not.


\paragraph{Results}
Table~\ref{tab:entity-search-results} reports for each resource the average number of results and their precision. We find that Google performs worst mainly because its diversification is limited (returns distinct answers often only far down in the ranking), and because it cannot well process conjunction and negation. Wikia performs better in terms of answer size, as by design it contains each entity only once. Still, it struggles with logical connectors. The Wikia categories produce more results than TiFi (9 vs.\ 6 on average), though due noise, they yield a substantially lower precision (-24\%). This corresponds to the core of the TiFi approach, which in step 1 and 2 is cleaning, i.e., leads to a lower recall while increasing precision.

Table 12
lists three sample queries along with their output. Crossed-out entities are incorrect answers. As one can see, text search mostly fails in answering the queries that use boolean connectives, while the original Wikia categories are competitive in terms of the number of answers, but produce many more wrong answers.

\begin{table}
\centering
\scalebox{1}{
\begin{tabular}{c|r|r|r|r}
\hline
\textbf{}              & \multicolumn{2}{c|}{\textbf{Text}}                                         & \multicolumn{2}{c}{\textbf{Structured Sources}}                                  \\ \hline
\textbf{Query}         & \multicolumn{1}{c|}{\textbf{Google}} & \multicolumn{1}{c|}{\textbf{Wikia}} & \multicolumn{1}{c|}{\textbf{Wikia-categories}} & \multicolumn{1}{c}{\textbf{TiFi}} \\ \hline
\multicolumn{1}{c|}{$t$} & 2 (52\%)                              & 7 (65\%)                             & 10 (62\%)                                     & 8 (87\%)                            \\ \hline
\multicolumn{1}{c|}{$t_1 \cap t_2$} & 1 (23\%)                              & 2 (11\%)                             & 8 (40\%)                                      & 3 (70\%)                            \\ \hline
\multicolumn{1}{c|}{$t_1 \setminus t_2$} & 1 (20\%)                              & 4 (36\%)                             & 8 (63\%)                                      & 6 (79\%)                            \\ \hline
\textbf{Average}       & 1 (32\%)                              & 4 (37\%)                             & 9 (55\%)                                      & 6 (79\%)                            \\ \hline
\end{tabular}}
\caption{Avg. \#Answers and precision of entity search.} 
\label{tab:entity-search-results}
\vspace{-0.75cm}
\end{table}

\section{Conclusion}

In this paper we have introduced TiFi, a system for taxonomy induction for fictional domains. TiFi uses a three-step architecture with category cleaning, edge cleaning, and top-level construction, thus building holistic domain specific taxonomies that are consistently of higher quality than what the Wikipedia-oriented state-of-the-art could produce.

Unlike most previous work, our approach is not based on static rules, but uses supervised learning. This comes with the advantage of allowing to rank classes and edges, for instance, in order to distinguish between core elements, less or marginally relevant ones, and totally irrelevant ones. In turn it also necessitates the generation of training data, yet we have shown that training data can be reasonably reused across domains.

Mirroring earlier experiences of YAGO~\cite{yago}, it also turns out that a crucial step in building useful taxonomies is the incorporation of abstract classes. For TiFi we relied on the established WordNet hierarchy, nevertheless finding the need to adapt a few links, and to remove certain too abstract concepts.

So far we only applied our system to fictional domains and one slice of Wikipedia. In the future, we would like to explore the construction of more  domain-specific but real-world taxonomies, such as gardening, Maya culture or Formula 1 racing.

Code and taxonomies will be made available on Github.



\bibliographystyle{ACM-Reference-Format}
\bibliography{references}  



\end{document}